\documentclass[a4paper]{article}
\usepackage[latin1]{inputenc} 
\usepackage[T1]{fontenc} 
\usepackage{RR,RRthemes}
\usepackage{hyperref}
\usepackage{amsmath,amssymb}

\def\y{{\mathbf y}}
\def\q{{\mathbf q}}
\def\x{{\mathbf x}}
\def\v{{\mathbf v}}
\def\a{{\mathbf a}}
\def\b{{\mathbf b}}
\def\Y{{\mathcal Y}}
\def\I{{\mathcal I}}

\def\BB{{\mathbb B}}
\def\real{{\mathbb R}}

\def\nub{\boldsymbol{\nu}}
\def\mub{\boldsymbol{\mu}}
\def\sign#1{\text{sign}(#1)}
\def\NN{{\textrm{NN}}}

\newenvironment{itemizerv}{
\begin{itemize}
}{\end{itemize}}

\RRNo{7771}
\RRdate{October 2011}

\RRauthor{
Herv\'e J\'egou \and Teddy Furon \and Jean-Jacques Fuchs
}
\authorhead{J\'egou \& Furon \& Fuchs}

\RRetitle{Anti-sparse coding \\ for approximate nearest neighbor search}
\RRtitle{Codage anti-parcimonieux pour la recherche approximative 
de plus proches voisins}
\titlehead{Anti-sparse coding for approximate search}

\RRnote{This work was realized as part of the Quaero Project, funded by OSEO, French State agency for innovation.}
\RRresume{Cet article proposes une technique de binarisation qui s'appuie sur le concept récent de \emph{codage anti-parcimonieux}, et montre ses excellentes performances dans un contexte de recherche approximative de plus proches voisins. Contrairement aux méthodes concurrentes, le cadre proposé permet, à un facteur d'échelle près, la reconstruction \emph{explicite} du vecteur encodé à partir de sa représentation binaire. L'article montre également que les projections aléatoires qui sont communément utilisées dans les méthodes de hachage multi-dimensionnel peuvent être avantageusement remplacées par des frames régulières lorsque le nombre de bits excède la dimension originale du descripteur. 
}
\RRabstract{This paper proposes a binarization scheme for vectors of high dimension based on the recent concept of \emph{anti-sparse} coding, and shows its excellent performance for approximate nearest neighbor search. Unlike other binarization schemes, this framework allows, up to a scaling factor, the \emph{explicit} reconstruction from the binary representation of the original vector. The paper also shows that random projections which are used in Locality Sensitive Hashing algorithms, are significantly outperformed by regular frames for both synthetic and real data if the number of bits exceeds the vector dimensionality, i.e., when high precision is required.}
\RRmotcle{codage parcimonieux, représentations étalées, recherche approximative de plus proches voisins, binarisation}
\RRkeyword{sparse coding, spread representations, approximate neighbors search, Hamming embedding}
\RRprojet{Texmex}  
\RRdomaineProj{texmex}

\RCRennes
\URRennes

\begin{document}
\makeRR   

\section{Introduction}
\label{sec:intro}

This paper addresses the problem of approximate nearest neighbor (ANN) search 
in high dimensional spaces.  
Given a query vector, the objective is to find, in a collection of vectors, 
those which are the closest to the query with respect to
a given distance function. We focus on the Euclidean distance 
in this paper. 
This problem has a very high practical interest, since matching the descriptors 
representing the media is the most consuming operation 
of most state-of-the-art audio~\cite{CVGLRS08}, image~\cite{JDS10a}
and video~\cite{LTJ07} indexing techniques.
There is a large body of literature on techniques whose aim is the optimization of the trade-off between retrieval time and complexity. 

We are interested by the techniques that regard
the memory usage of the index as a major criterion. This is compulsory 
when considering large datasets including dozen millions to billions 
of vectors~\cite{TFW08,WTF09,JDS10a,JTDA11}, 
because the indexed representation must fit in memory to avoid costly hard-drive accesses. 
One popular way is to use 
a Hamming Embedding function that maps the real vectors into binary vectors~\cite{TFW08,WTF09,JDS10a}: 
Binary vectors are compact, and searching the Hamming space is efficient (\texttt{XOR} 
operation and bit count) even if the comparison is exhaustive between the binary query and the database vectors. 
An extension to these techniques is the  asymmetric 
scheme~\cite{DCL08,JDS11} which limits the approximation done on the query, 
leading to better results for a slightly higher complexity. 

We propose to address the ANN search problem  with an \emph{anti-sparse} solution based on the design of \emph{spread} representations recently proposed by Fuchs~\cite{F11}. 
Sparse coding  has received in the last decade
a huge attention from both theoretical and practical points of view. 
Its objective is to represent a vector in a higher dimensional space with a very limited number of non-zeros components.
Anti-sparse coding has the opposite properties.
It offers a robust representation of a vector in a higher dimensional space with all the components sharing evenly the information.

 Sparse and anti-sparse coding admits a common formulation. The algorithm proposed by Fuchs~\cite{F11} 
is indeed similar to path-following methods based on continuation techniques like~\cite{MR2060166}.
The anti-sparse problem considers a $\ell_{\infty}$ penalization term 
where the sparse problem usually considers the $\ell_1$ norm.
The penalization in $\|\x\|_{\infty}$ limits the range of the coefficients 
which in turn tend to `stick' their value to $\pm\|\x\|_{\infty}$~\cite{F11}.
As a result, the anti-sparse approximation offers a natural binarization method.
 
Most importantly and in contrast to other Hamming Embedding techniques, 
the binarized vector allows an explicit and reliable reconstruction of the original 
database vector. This reconstruction is very useful to refine the search.
First, the comparison of the Hamming distances between the binary representations 
identifies some potential nearest neighbors. 
Second, this list is refined by computing the Euclidean distances between the query 
and the reconstructions of the database vectors.


We provide a Matlab package to reproduce the analysis comparisons reported in this paper (for the tests 
on synthetic data), see \url{http://www.irisa.fr/texmex/people/jegou/src.php}. 
The paper is organized as follows. Section~\ref{sec:spread} introduces the anti-sparse coding framework.
Section~\ref{sec:indexing} describes the corresponding ANN search method 
which is evaluated in Section~\ref{sec:experiments} on both synthetic and real data.

\section{Spread representations}
\label{sec:spread}

This section briefly describes the anti-sparse coding of~\cite{F11}.
We first introduce the objective function and provide the guidelines of the algorithm
giving the spread representation of a given input real vector.

Let $A=[\a_{1}|\ldots|\a_{m}]$ be a $d\times m$ ($d<m$) full rank matrix.
For any $\y\in\real^{d}$, the system $A\x=\y$ admits an infinite number of solutions.
To single out a unique solution, one add a constraint as for instance seeking a minimal norm solution.
Whereas the case of the Euclidean norm is trivial, and the case of the $\ell_{1}$-norm stems in the vast literature of sparse representation, Fuchs recently studied the case of the $\ell_{\infty}$-norm. Formally, the problem is:
\begin{equation}
\x^{\star} =\min_{\x:\,A\x=\y} \|\x\|_{\infty},
\label{eq:MinInf}
\end{equation}
with $\|\x\|_{\infty} = \max_{i\in\{1,\ldots,m\}} |x_{i}|$. Interestingly, he proved that by minimizing the range of the components, $m-d+1$ of them are stuck to the limit, ie. $x_{i}=\pm \|\x\|_{\infty}$. Fuchs also exhibits an efficient way to solve~\eqref{eq:MinInf}. He proposes to solve the series of simpler problems 
\begin{equation}
\x^{\star}_{h} = \min_{\x\in\real^{m}} J_{h}(\x)
\end{equation} 
with 
\begin{equation}
J_{h}(\x)=\|A\x-\y\|_{2}^{2}/2 + h\|\x\|_{\infty}
\end{equation}
for some decreasing values of $h$. As $h\rightarrow0$,   $\x^{\star}_{h}\rightarrow \x^{\star}$.
\subsection{The sub-differential set}
For a fixed $h$, $J_{h}$ is not differentiable due to $\|.\|_{\infty}$. Therefore,  we need to work with sub-differential sets.
The sub-differential set $\partial f(\x)$ of function $f$ at $\x$ is the set of gradients $\v$ s.t. $f(\x')-f(\x)\geq \v^{\top}(\x'-\x),\,\forall \x'\in\real^{m}$.
For $f\equiv\|.\|_{\infty}$, we have:
\begin{eqnarray}
\partial f(\mathbf{0})&=&\{\v\in\real^{m}: \|\v\|_{1}\leq1\},\label{eq:SubDiff0}\\
\partial f(\x)&=&\{\v\in\real^{m}: \|\v\|_{1}=1,\label{eq:SubDiffx}\\
&&v_{i}x_{i}\geq 0\text{ if } |x_{i}|=\|\x\|_{\infty},\nonumber\\
&&v_{i}=0\text{ else}\nonumber\}\,\text{, for } \x\neq\mathbf{0}
\end{eqnarray}
Since $J_{h}$ is convex, $\x^{\star}_{h}$ is solution iff $\mathbf{0}$ belongs to the sub-differential set $\partial J_{h}(\x^{\star}_{h})$, i.e. iff there exist $\v\in\partial f(\x^{\star}_{h})$ s.t.
\begin{equation}
A^{\top}(A\x^{\star}_{h}-\y)+h\v=\mathbf{0}
\label{eq:Deriv}
\end{equation}

\subsection{Initialization and first iteration}
\label{sub:Init} 
For $h_{0}$ large enough, $J_{h_{0}}(\x)$ is dominated by $\|\x\|_{\infty}$, and the solution writes $\x^{\star}_{h_{0}}=\mathbf{0}$ and $\v=h_{0}^{-1}A^{\top}\y\in\partial f(\mathbf{0})$. \eqref{eq:SubDiff0} shows that this solution no longer holds for $h<h_{1}$ with $h_{1}=\|A^{\top}\y\|_{1}$.

For $\|\x\|_{\infty}$ small enough, $J_{h}(\x)$ is dominated by $\|\y\|^{2}-\x^{\top}A^{\top}\y+h\|\x\|_{\infty}$ whose minimizer is $\x^{\star}_{h}=\|\x\|_{\infty}\sign{A^{\top}\y}$. In this case, $\partial f(\x)$ is the set of vectors $\v$ s.t. $\sign{\v}=\sign{\x}$ and $\|\v\|_{1}=1$. Multiplying~\eqref{eq:Deriv} by $\sign{\v}^{\top}$ on the left, we have
\begin{equation}
h = h_{1}-\|A\sign{A^{\top}\y}\|^{2}\|\x\|_{\infty}.
\label{eq:h_xinf}
\end{equation}
This shows that i) $\x^{\star}_{h}$ can be a solution for $h<h_{1}$, and ii) $\|\x\|_{\infty}$ increases as $h$ decreases. Yet, Equation \eqref{eq:Deriv} also imposes that $\v = \nub_{1}-\mub_{1}\|\x\|_{\infty}$, with 
\begin{equation}
\nub_{1}\triangleq h^{-1}A^{\top}\y \hspace{1cm} \text{and} \hspace{1cm} \mub_{1}\triangleq h^{-1}A^{\top}A\sign{A^{\top}\y}.
\end{equation}
 But, the condition $\sign{\v}=\sign{\x}$ from~\eqref{eq:SubDiffx} must hold. This limits $\|\x\|_{\infty}$ by $\rho_{i_{2}}$ where $\rho_{i} = \nu_{i}/\mu_{i}$ and $i_{2}=\arg\min_{i:\rho_{i}>0}(\rho_{i})$, which in turn translates to a lower bound $h_{2}$ on $h$ via~\eqref{eq:h_xinf}. 

\subsection{Index partition}
For the sake of simplicity, we introduce $\I\triangleq\{1,\ldots,m\}$, and the index partition $\bar{\I}\triangleq\{i:|x_{i}|=\|\x\|_{\infty}\}$ and $\breve{\I}\triangleq\I\setminus\bar{\I}$. The restriction of vectors and matrices to $\bar{\I}$ (resp. $\breve{\I}$) are denoted alike $\bar{\x}$ (resp. $\breve{\x}$). For instance, Equation~\eqref{eq:SubDiffx} translates in $\sign{\bar{\v}}=\sign{\bar{\x}}$, $\|\bar{\v}\|_{1}=1$ and $\breve{\v}=\mathbf{0}$.
The index partition splits~\eqref{eq:Deriv} into two parts:
\begin{eqnarray}
\breve{A}^{\top}\left(\breve{A}\breve{\x}+\bar{A}\sign{\bar{\v}}\|\x\|_{\infty}\right)=\breve{A}^{\top}\y\label{eq:Deriv1}\\
\bar{A}^{\top}\left(\breve{A}\breve{\x}+\bar{A}\sign{\bar{\v}}\|\x\|_{\infty}-\y\right)=-h\bar{\v}\label{eq:Deriv2}
\end{eqnarray}
For $h_{2}\leq h< h_{1}$, we've seen that $\bar{\x}=\x$, $\bar{\v}=\v$, and $\bar{A}=A$. Their `tilde' versions are empty. For $h<h_{2}$, the index partition $\bar{\I}=\I$ and $\breve{\I}=\emptyset$ can no longer hold. Indeed, when $v_{i_{2}}$ is null at $h=h_{2}$, the $i_{2}$-th column of $A$ moves from $\bar{A}$ to $\breve{A}$ s.t. now, $\breve{A}=[\a_{i_{2}}]$.

\subsection{General iteration}
The general iteration consists in determining on which interval $[h_{k+1},h_{k}]$ an index partition holds, giving the expression of the solution $\x^{\star}_{h}$ and proposing a new index partition to the next iteration.

Provided $\breve{A}$ is full rank, \eqref{eq:Deriv1} gives
\begin{equation}
\breve{\x}=\boldsymbol{\xi}_{k} +  \boldsymbol{\zeta}_{k} \|\x\|_{\infty},\label{eq:tildex}
\end{equation}
with
\begin{equation}
\boldsymbol{\xi}_{k}=(\breve{A}^{\top}\breve{A})^{-1}\breve{A}^{\top}\y
\end{equation} 
and 
\begin{equation}
\boldsymbol{\zeta}_{k}= - (\breve{A}^{\top}\breve{A})^{-1}\bar{A}\sign{\bar{\v}}.
\end{equation}
Equation~\ref{eq:Deriv2} gives:
\begin{equation}
\bar{\v}=\nub_{k}-\mub_{k}\|\x\|_{\infty},\quad\label{eq:barv}
\end{equation}
with
\begin{equation}
\mub_{k}=\bar{A}^{\top}(I-\bar{A}^{\top}\breve{A}(\breve{A}^{\top}\breve{A})^{-1})\bar{A}\sign{\bar{\v}}/h
\end{equation} 
and 
\begin{equation}
\nub_{k}=(\breve{A}^{\top}\y - \boldsymbol{\xi}_{k})/h.
\end{equation} 
Left multiplying \eqref{eq:Deriv2} by $\sign{\bar{\v}}$, we get:
\begin{equation}
h = \eta_{k} -\upsilon_{k}\|\x\|_{\infty}\label{eq:h}
\end{equation}
with 
\begin{equation}
\upsilon_{k}=(\bar{A}\sign{\bar{\v}})^{\top}\left(I-\breve{A}(\breve{A}^{\top}\breve{A})^{-1}\breve{A}^{\top}\right)\bar{A}\sign{\bar{\v}}, 
\end{equation}
and
\begin{equation}
\eta_{k}=-\sign{\bar{\v}}^{\top}\bar{A}^{\top}(\breve{A}\breve{\x}-\y).
\end{equation}
Note that $\upsilon_{k}>0$ so that $\|\x\|_{\infty}$ increases when $h$ decreases. 

These equations extend a solution $\x^{\star}_{h}$ to the neighborhood of $h$.
However, we must check that this index partition is still valid as we decrease $h$ and $\|\x\|_{\infty}$ increases.
Two events can break the validity:
\begin{itemizerv}
\item Like in the first iteration, a component of
$\bar{\v}$ given in~\eqref{eq:barv} becomes null. This index moves from $\bar{\I}$ to $\breve{\I}$.
\item A component of $\breve{\x}$ given in~\eqref{eq:tildex} sees its amplitude equalling $\pm\|\x\|_{\infty}$.
This index moves from $\breve{\I}$ to $\bar{\I}$, and the sign of this component will be the sign of the new component of $\bar{\x}$.
\end{itemizerv}
The value of $\|\x\|_{\infty}$ for which one of these two events first happens is translated in $h_{k+1}$ thanks to~\eqref{eq:h}.

\subsection{Stopping condition and output}
If the goal is to minimize $J_{h_{t}}(\x)$ for a specific target $h_{t}$, then the algorithm stops when $h_{k+1}<h_{t}$.
The real value of $\|\x^{\star}_{h_{t}}\|_{\infty}$ is given by~\eqref{eq:h}, and the components not stuck to $\pm\|\x^{\star}_{h_{t}}\|_{\infty}$ 
by~\eqref{eq:tildex}.

We obtain the spread representation~$\x$ of the input vector~$\y$. The vector $\x$ has many of its components equal 
to $\pm\|\x\|_{\infty}$. An approximation of the original vector~$\y$ is obtained by
\begin{equation}
\hat{\y} = A\x.
\label{equ:reconstruction}
\end{equation}

\section{Indexing and search mechanisms}
\label{sec:indexing}
This section describes how Hamming Embedding functions are used 
for approximate search, and in particular how the anti-sparse coding framework described 
in Section~\ref{sec:spread} is exploited. 

\subsection{Problem statement}
Let $\Y$ be a dataset of $n$ real vectors, $\Y=\{\y_1,\dots,\y_n\}$, 
where $\y_i \in \real^d$, and consider a query vector $\q \in \real^d$. 
We aim at finding the $k$ vectors in~$\Y$ that are closest 
to the query, with respect to the Euclidean distance. For the sake of exposure, 
we consider without loss of generality the nearest neighbor problem, i.e., the case $k=1$. 
The nearest neighbor of $\q$  in $\Y$ is defined as 
\begin{equation}
\NN(\q) = \arg \min_{\y \in \Y} \|\q - \y\|^2.
\end{equation}

The goal of approximate search is to find this nearest neighbor with high probability 
and using as less resources as possible. The performance criteria 
are the following:
\begin{itemizerv}
\item The quality of the search, i.e., to which extent the algorithm is able to return 
the true nearest neighbor ; 
\item The search efficiency, typically measured by the query time ;
\item The memory usage, i.e., the number of bytes used to index a vector~$\y_i$ of the database. 
\end{itemizerv}
In our paper, we assess the search quality by the recall@R measure: over 
a set of queries, we compute the proportion for which the system returns the true 
nearest neighbor in the first R positions. 

\subsection{Approximate search with binary embeddings}
A class of ANN methods is based on embedding~\cite{TFW08,WTF09,JDS10a}. 
The idea is to map the input vectors to a space where the representation is compact and the comparison is efficient.
The Hamming space offers these two desirable properties.
The key problem is the design of the embedding 
function $e:\real^d\rightarrow \BB^m$ mapping the input vector $\y$ to $\b = e(\y)$ in the $m$-dimensional Hamming space $\BB^m$, 
here defined as $\{-1,1\}^m$ for the sake of exposure. 

Once this function is defined, all the database vectors are mapped to $\BB^m$, 
and the search problem is translated into the Hamming space based on the Hamming distance, 
or, equivalently:
\begin{equation}
\NN_b\left(e(\q)\right) = \arg \max_{\y \in \Y} e(\q)^{\top} \, e(\y). 
\label{equ:ip1}
\end{equation}
$\NN_{b}(e(\q))$ is returned as the approximate $\NN(\q)$.

\medskip
{\noindent \bf Binarization with anti-sparse coding.}
Given an input vector~$\y$, the anti-sparse coding of Section~\ref{sec:spread} 
produces $\x$ with many components equal to $\pm ||x||_{\infty}$. 
We consider a ``pre-binarized'' version $\dot{x}(\y)=\x/\|\x\|_{\infty}$, 
and the binarized version $e(\y)=\sign{\x}$. 

\subsection{Hash function design}
\label{sec:frames}
The locality sensitive hashing (LSH) algorithm is mainly based on random projection, though 
different kinds of hash functions have been proposed for the Euclidean space~\cite{AnI06}.
Let $A=[\a_{1}|\ldots|\a_{m}]$ be a $d\times m$ matrix storing the $m$ projection vectors.
The most simple way is to take the sign of the projections: $\b=\sign{A^{\top}\y}$.
Note that this corresponds to the first iteration of our algorithm (see Section~\ref{sub:Init}).

We also try $A$ as an uniform frame. 
A possible construction of such a frame consists in 
performing a QR decomposition on a $m\times m$ matrix. The matrix 
$A$ is then composed of the $d$ first rows of the $Q$ matrix, 
ensuring that $A \times A^{\top} = {\mathbb I}_d$. 
Section~\ref{sec:experiments} shows that such frames 
significantly improve the results compared with random projections, 
for both LSH and anti-sparse coding embedding methods. 

\subsection{Asymmetric schemes}
\label{sec:asymmetric}
As recently suggested 
in the literature, a better search quality is obtained by avoiding the binarization 
of the query vector. Several variants are possible. We consider 
the simplest one derived from~(\ref{equ:ip1}), where the query is not binarized 
in the inner product. For our anti-sparse coding scheme, 
this amounts to performing the search based on the following maximization:
\begin{equation}
\NN_{a}\left(e(\q)\right) = \arg \max_{\y \in \Y} \dot{x}(\q)^{\top} e(\y).
\label{equ:ip2}
\end{equation}
The estimate $\NN_a$ is better than $\NN_b$. The memory usage is the same 
because the vectors in the database $\{e(\y_{i})\}$ are all binarized.  
However, this asymmetric scheme is a bit slower than the pure bit-based comparison.  
For better efficiency, the search~\eqref{equ:ip2} is done using look-up 
tables computed for the query and prior to the comparisons~\cite{JDS11}.
This is slightly slower than computing the Hamming distances in~\eqref{equ:ip1}.
This asymmetric scheme is interesting for any binarization scheme (LSH or anti-sparse coding) 
and any definition of $A$ (either random projections or a frame). 

\subsection{Explicit reconstruction}
\label{sec:explicit}
The anti-sparse binarization scheme explicitly minimizes the reconstruction error, 
which is traded in~(\ref{eq:MinInf}) with the $\ell_{\infty}$ regularization term.
Equation~\eqref{equ:reconstruction} gives an explicit approximation of the database vector $\y$ 
up to a scaling factor: $\hat{\y} \propto \frac{A \b}{||A \b||_2}$.
The approximate nearest neighbors $\NN_e$ are obtained by computing the exact Euclidean distances 
$||\q - \hat{\y}_{i}||_2$. 
This is slow compared to the Hamming distance computation. 
That is why, it is used to operate, like in~\cite{JTDA11}, a re-ranking of the first hypotheses 
returned based on the Hamming distance (on the asymmetric 
scheme described in Section~\ref{sec:asymmetric}). The main difference with~\cite{JTDA11} 
is that no extra-code has to be retrieved: the reconstruction $\hat{\y}$ solely relies on $\b$. 



\section{Simulations and experiments}
\label{sec:experiments}
This section evaluates the search quality on synthetic and real data. 
In particular, we measure the impact of:
\begin{itemizerv}
\item The Hamming embedding technique: LSH and binarization based on anti-sparse coding. 
We also compare to the spectral hashing method of~\cite{WTF09}, using the code 
available online. 
\item The choice of matrix~$A$: random projections or frame for LSH. 
For the anti-sparse coding, we always assume a frame. 
\item The search method: 1) $\NN_b$ of~\eqref{equ:ip1} 2) $\NN_a$ of~\eqref{equ:ip2} and 3) $\NN_e$ 
as described in Section~\ref{sec:explicit}. 
\end{itemizerv}

Our comparison focuses on the case $m \geq d$.  
In the anti-sparse coding method, the regularization term~$h$ controls 
the trade-off between the robustness of the Hamming embedding and the quality 
of the reconstruction. Small values of $h$ favors the quality of the reconstruction 
(without any binarization). Bigger values 
of $h$ gives more components stuck to $\|\x\|_{\infty}$, 
which improves the approximation search with binary embedding. 
Optimally, this parameter should be adjusted to 
give a reasonable trade-off between the efficiency of
the first stage (methods $\NN_b$ or $\NN_a$) and the re-ranking stage ($\NN_e$). 
Note however that, thanks to the algorithm described in Section~\ref{sec:spread}, 
the parameter is stable, i.e., a slight modification of this parameter only affects a few components.
We set $h=1$ in all our experiments. 
Two datasets are considered for the evaluation:
\begin{itemizerv}
\item A database of 10,000 16-dimensional vectors uniformly drawn on the 
Euclidean unit sphere (normalized Gaussian vectors) and a set of 1,000 query vectors. 
\item A database of SIFT~\cite{Low04} descriptors available 
online\footnote{http://corpus-texmex.irisa.fr}, 
comprising 1 million database and 10,000 query vectors of dimensionality 128. 
Similar to~\cite{WTF09}, we first reduce the vector dimensionality to 48 
components using principal component analysis (PCA). 
The vectors are not normalized after PCA. 
\end{itemizerv}

\begin{figure}
\includegraphics[width=\linewidth]{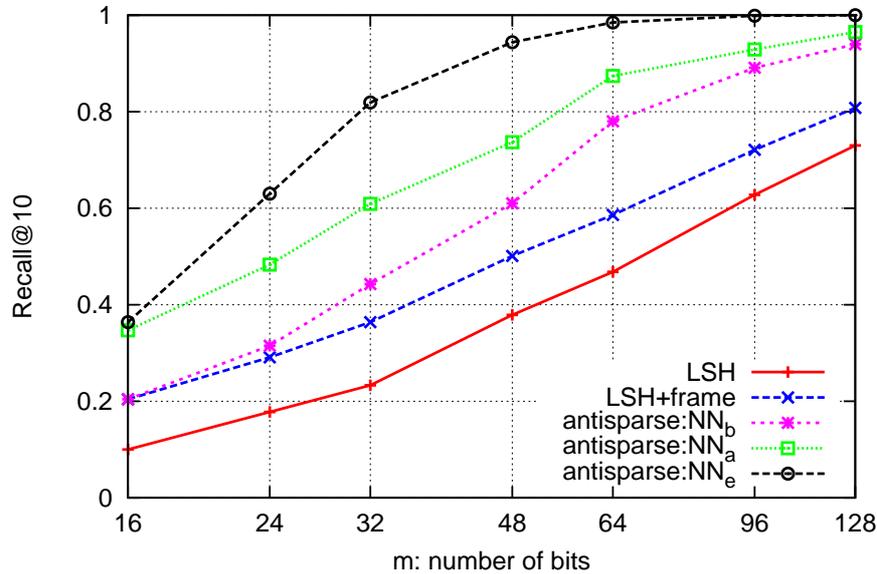}
\caption{Anti-sparse coding \emph{vs} LSH on synthetic data.
Search quality (recall@10 in a vector set of 10,000 vectors) as a function of the number of bits of the representation.} 
\label{fig:nbbits}
\end{figure}

\begin{figure}
\includegraphics[width=\linewidth]{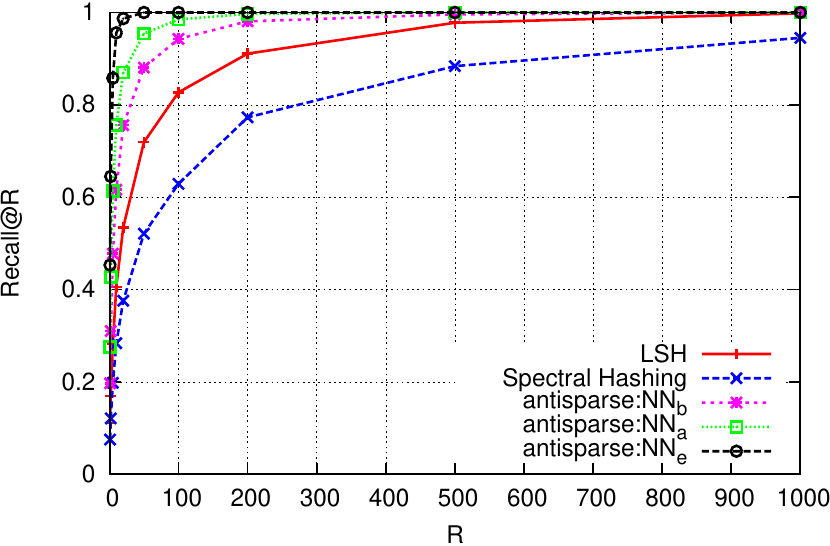}
\caption{Anti-sparse coding \emph{vs} LSH on synthetic data ($m=48$, 10,000 vectors in dataset). 
\label{fig:lshvsas}}
\end{figure}

{\noindent \bf The comparison of LSH and anti-sparse.} Figures~\ref{fig:nbbits} and~\ref{fig:lshvsas}
show the performance of Hamming embeddings for synthetic data. 
On Fig.~\ref{fig:nbbits}, the quality measure is the recall@10 (proportion of true NN
ranked in first 10 positions) plotted as a function of the number of bits~$m$. 
For LSH, observe the much better performance obtained by the proposed frame construction 
compared with random projections. 
The same conclusion holds for anti-sparse binarization. 

The anti-sparse coding offers similar search quality as LSH for $m=d$ 
when the comparison is performed using  $\NN_b$ of~\eqref{equ:ip1}. 
The improvement gets significant as $m$ increases. 
The spectral hashing technique~\cite{WTF09} exhibits poor performance on this 
synthetic dataset. 
\medskip

{\noindent \bf The asymmetric comparison $\NN_a$} leads a significant improvement, 
as already observed in~\cite{DCL08,JDS11}. 
The interest of anti-sparse coding becomes obvious by considering the performance 
of the comparison~$\NN_e$ based on the explicit reconstruction of the database 
vectors from their binary-coded representations. 
For a fixed number of bits, the improvement is huge compared to
LSH. It is worth using this technique to re-rank the first hypotheses 
obtained by $\NN_b$ or $\NN_a$.
\medskip

\begin{figure}
\includegraphics[width=\linewidth]{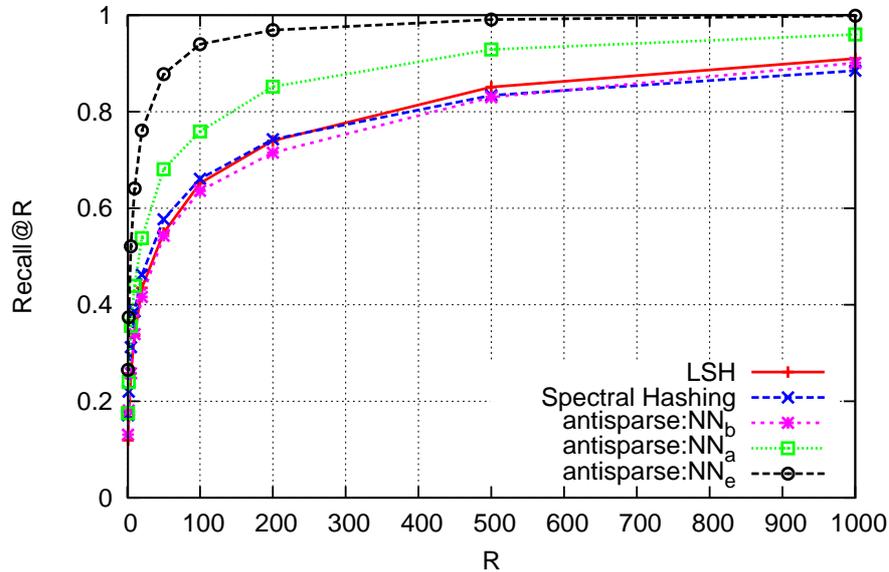}
\caption{Approximate search in a SIFT vector set of 1 million vectors. 
\label{fig:sift}}
\end{figure}

{\noindent \bf Experiments on SIFT descriptors.} As shown by Figure~\ref{fig:sift}, 
LSH is slightly better than anti-sparse on real data 
when using the binary representation only (here $m=128$), 
which might solved by tuning $h$, 
since the first iteration of antisparse leads the binarization as LSH. 
However, the interest of the explicit reconstruction offered by $\NN_e$ is again obvious. 
The final search quality is significantly better than that obtained by spectral 
hashing~\cite{WTF09}. 
Since we do not specifically handle the fact that our descriptor are not normalized after PCA, 
our results could probably be improved by taking care of the $\ell_2$ norm. 

\section{Conclusion and open issues}

In this paper, we have proposed anti-sparse coding as an effective Hamming embedding,
which, unlike concurrent techniques, offers an explicit reconstruction 
of the database vectors. To our knowledge, it outperforms all 
other search techniques based on binarization. 
There are still two open issues to take the best of the method. 
First, the computational cost is still a bit high for high dimensional vectors.
Second, if the proposed codebook construction is better than random projections, 
it is not yet specifically adapted to real data. 


 \newpage
\tableofcontents

\end{document}